\documentclass{article}

\usepackage{microtype}
\usepackage{graphicx}
\usepackage{subfigure}
\usepackage{booktabs} 

\usepackage{xcolor}
\usepackage{makecell}
\usepackage{enumitem}
\usepackage{xspace}
\usepackage{adjustbox}

\usepackage{hyperref}

\newcommand{\methodname}{Speculative Streaming}

\usepackage[accepted]{icmlarxiv}

\usepackage{amsmath}
\usepackage{amssymb}
\usepackage{mathtools}
\usepackage{amsthm}
\usepackage{multirow}

\usepackage[capitalize,noabbrev]{cleveref}

\theoremstyle{plain}

\theoremstyle{definition}

\theoremstyle{remark}

\usepackage[textsize=tiny]{todonotes}

\icmltitlerunning{Speculative Streaming: Fast LLM Inference without Auxiliary Models}

\begin{document}
\twocolumn[
\icmltitle{Speculative Streaming: Fast LLM Inference without Auxiliary Models}

\icmlsetsymbol{equal}{*}

\begin{icmlauthorlist}
\icmlauthor{Nikhil Bhendawade}{apple}
\icmlauthor{Irina Belousova}{apple}
\icmlauthor{Qichen Fu}{apple}
\icmlauthor{Henry Mason}{apple}
\icmlauthor{Mohammad Rastegari}{apple}
\icmlauthor{Mahyar Najibi}{apple}
\end{icmlauthorlist}
\icmlcorrespondingauthor{Nikhil Bhendawade}{nbhendawade@apple.com}
\icmlcorrespondingauthor{Irina Belousova}{ibelousova@apple.com}
\icmlcorrespondingauthor{Qichen Fu}{qfu22@apple.com}
\icmlcorrespondingauthor{Henry Mason}{hmason@apple.com}
\icmlcorrespondingauthor{Mohammad Rastegari}{mrastegari@apple.com}
\icmlcorrespondingauthor{Mahyar Najibi}{najibi@apple.com}
\icmlaffiliation{apple}{Apple}

\icmlkeywords{Large Language Models, Speculative Decoding}

\vskip 0.3in
]

\printAffiliationsAndNotice{} 

\begin{abstract}
Speculative decoding is a prominent technique to speed up the inference of a large target language model based on predictions of an auxiliary draft model. While effective, in application-specific settings, it often involves fine-tuning both draft and target models to achieve high acceptance rates. As the number of downstream tasks grows, these draft models add significant complexity to inference systems. We propose \methodname\xspace, a single-model speculative decoding method that fuses drafting into the target model by changing the fine-tuning objective from next token prediction to future n-gram prediction. \methodname\xspace speeds up decoding by 1.8 - 3.1X in a diverse set of tasks, such as Summarization, Structured Queries, and Meaning Representation, without sacrificing generation quality. Additionally, \methodname\xspace is parameter-efficient. It achieves on-par/higher speed-ups than Medusa-style architectures while using $\sim$10000X fewer extra parameters, making it well-suited for resource-constrained devices.
\end{abstract}

\begin{figure}
    \centering
    \includegraphics[width=\linewidth]{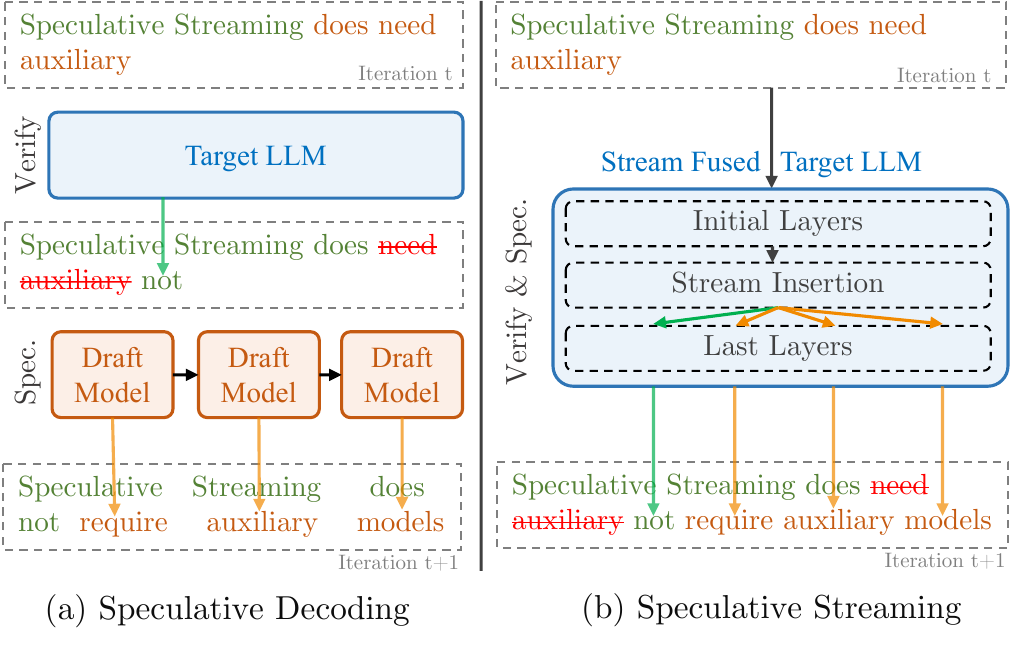}
    \caption{(a) Speculative Decoding requires a separate draft model that runs autoregressively to speculate. (b) Speculative Streaming significantly simplifies the system by performing speculation and verification concurrently, all within a single stream-fused model.}
    \label{fig:teaser}
\end{figure}

\section{Introduction}
\label{sec:intro}

Large transformers are today's preeminent tool for language modeling. The quality of these models improves as they scale~\cite{kaplan2020scaling}, leading to the introduction of the state-of-the-art multi-billion parameter models~\cite{brown2020language,thoppilan2022lamda,chowdhery2023palm,touvron2023llama}. While these models are very effective for token generation, they incur a high inference cost as the model and its transient states need to be loaded into computing memory for each subsequently generated token. Moreover, scaling up these models, besides making each call more compute-intensive, also makes their autoregressive generation memory bound~\cite{pope2023efficiently}, preventing them from making effective use of available compute. This poses a significant challenge to the deployment of large autoregressive transformers,
particularly for user-facing applications with tight latency requirements. 

Given the memory-bound nature of large language model (LLM) inference, recent work~\cite{leviathan2023fast, chen2023accelerating} proposed Speculative Decoding as an effective technique to accelerate decoding based on concepts borrowed from speculative computation~\cite{burton1985speculative} to exploit the available extra compute.
The core of speculative decoding is the idea of speculating multiple candidate future tokens first, and then verifying them all in parallel. To achieve this, as shown in \cref{fig:teaser}(a), a two-model paradigm approach is used: a small auxiliary ``draft'' model for candidate speculation and a large ``target'' model for verification~\cite{leviathan2023fast, chen2023accelerating}.
Although effective in accelerating LLMs, speculative decoding complicates deployment. Training also becomes more demanding and complicated, as a separate draft model needs to be trained and aligned with the target model. It is also not resource-efficient, requiring to host two models in memory during token prediction. This increased footprint is especially unsatisfactory for resource-constrained devices.

In this paper, we propose \textit{\methodname}, a single-model speculative decoding approach that unifies speculation and verification, obviating the need for a separate draft model as shown in \cref{fig:teaser}(b). This is accomplished by incorporating multi-stream attention into the target model to perform n-gram prediction which serves as future candidate speculation. As a result, a forward model pass can verify the previously generated tokens while simultaneously speculating on the future tokens. Moreover, compared to previous approaches, \methodname\xspace is trained end-to-end, naturally aligning speculation and verification phases.

While making the system significantly simpler and resource efficient, \methodname\xspace achieves speedups comparable to two-stage speculative decoding~\cite{leviathan2023fast} without degrading the quality metrics on a diverse set of downstream tasks. It also leads to on-par or better speedup than the more recent block-wise decoding model, Medusa~\cite{medusa}, that introduces multiple additional high-dimensional prediction heads. Moreover, \methodname\xspace requires 10000X fewer additional parameters than Medusa\cite{medusa}, which makes it an ideal method for resource-constrained devices.

In summary, the advantages of \methodname\xspace are as follows:
\begin{itemize}[leftmargin=*]
\vspace{-2mm}
\setlength\itemsep{0em}
\item[--] Achieving substantial speedups through streamlined fine-tuning and eliminating the need for a separate draft model. 
\item[--] Demonstrating resource efficiency with 10000X fewer extra parameters compared to ~\cite{medusa} while achieving better speedups, all without compromising quality across a diverse range of tasks.
\item[--] Simplifying deployment processes by eliminating the need to manage, align, and switch between two models during execution, as observed in~\cite{leviathan2023fast}.
\end{itemize}
\section{Related Works}
\label{sec:related}

The inference of large language models is often limited by the sequential nature of auto-regressive decoding, where each token generation requires a complete network forward pass. Several approaches have been proposed to address the high inference latency of large language models by directly decreasing the memory footprint of LLMs. Model quantization~\cite{frantar2022gptq,yao2022zeroquant,dettmers2023spqr}, knowledge distillation to a smaller a model~\cite{gu2023knowledge,agarwal2023gkd}, and pruning~\cite{frantar2023sparsegpt,sun2023simple} are among these techniques. 
Recently, speculative decoding (SD) has emerged as a vital technique to accelerate autoregressive decoding.

The original speculative decoding approach~\cite{chen2023accelerating, leviathan2023fast} utilizes a smaller 
LLM (\aka the \emph{draft model}), to generate a candidate sequence of tokens to be verified by the \emph{target model}.
With a well-tuned draft model, this technique can achieve a 2-3x inference speedup.
Recent SD variants propose parallel computation along the batch axis \cite{sun2023spectr}, and tree-structured batches \cite{miao2023specinfer, spector2023accelerating} to improve the acceptance rates of the guessed tokens by the target model and to further boost the performance.
However, these methods encounter a common limitation: the necessity of developing an accurate and independent draft model.
First, training such a draft model aligned with the main model is not trivial \cite{zhou2023distillspec}.
Second, hosting two different models increases the system complexity, and is more computationally and operationally expensive to maintain.

\begin{figure*}
    \centering
    \includegraphics[width=0.8\linewidth]{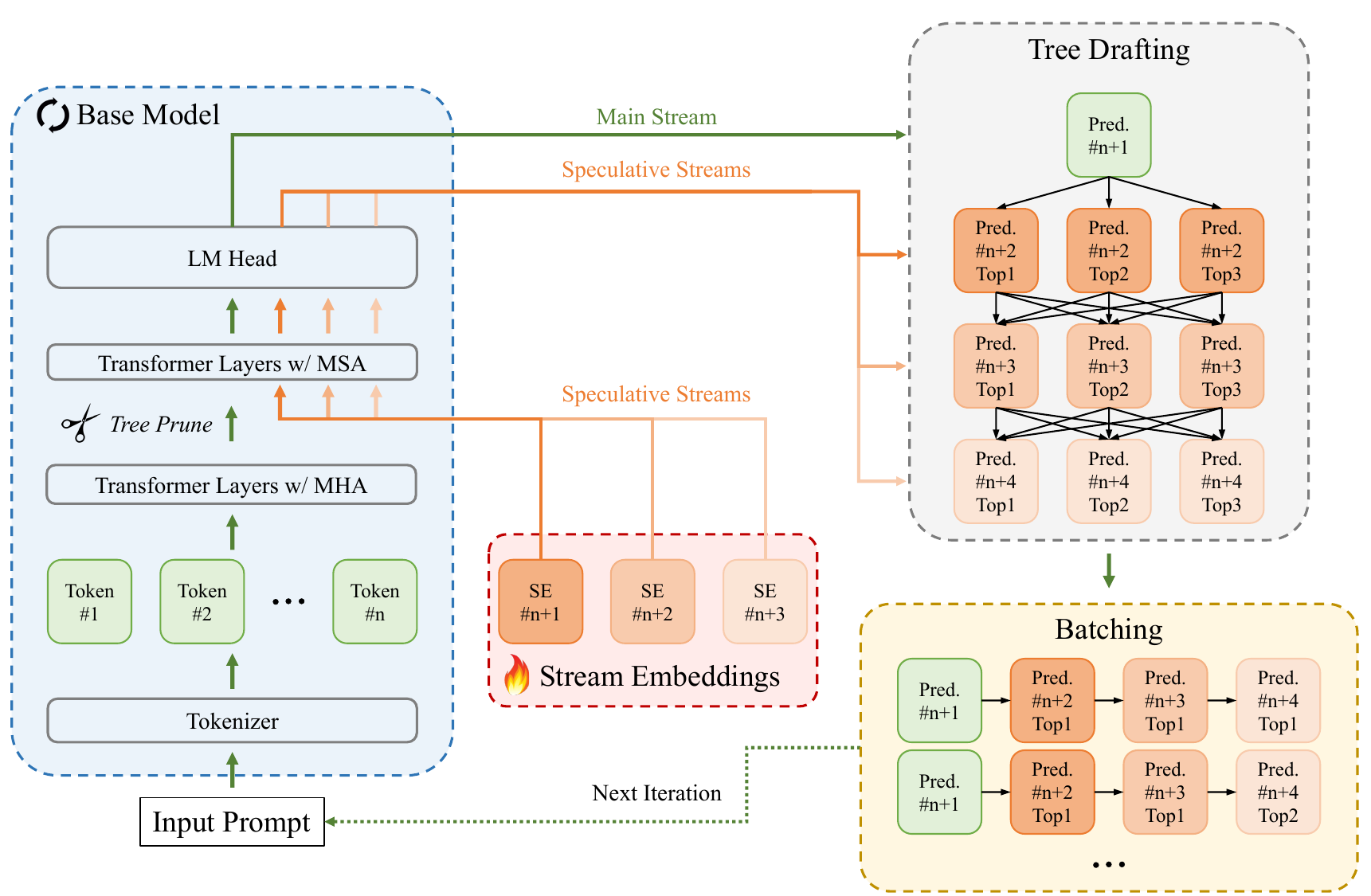}
    \caption{Architecture: We replace top $N_s$ multi-head attention (MHA) layers of the base model with multi-stream attention (MSA) layers as described in \eqref{eq:spec_stream_msa}. Speculative streams are initialized using hidden states of layer $N-N_s$ and stream identifier embeddings (SE), as described in \eqref{eq:spec_stream_init} and used to generate speculative draft in the form of a tree. The speculative tree draft from the previous iteration is batched for verification and pruned before stream insertion. During each forward pass previous tree draft is verified and a new tree draft is issued using speculative streams as described in \ref{parallel_spec}}
    \label{fig:main_fig}
\end{figure*}

Very recently, single-model speculation has also been considered.
In particular, inspired by~\cite{qi2020prophetnet,stern2018blockwise}, Medusa~\cite{medusa} extends the main model to predict future tokens in parallel by training multiple output heads.
While it does not require a draft model, each Medusa head of size (\emph{hidden\_size} $\times$ \emph{vocab\_size}) requires nonnegotiable additional parameters. As auto-regressive generation typically follows a memory-bound compute pattern, the significant amount of extra parameters introduces deployment challenges on resource-constrained devices. Alternatively, lookahead decoding~\cite{fu2023lookahead} proposes a parallel decoding method without learning new parameters.
It uses Jacobi Decoding and n-gram history trajectory cache to generate and verify future n-gram prediction. 
Differently, Speculative Streaming achieves n-gram generation by learning a set of token embeddings and accelerates decoding by running speculation and verification concurrently. Notably, our approach can achieve better speedup with significantly less number of extra parameters compared to the existing speculative decoding methods \cite{chen2023accelerating, leviathan2023fast, medusa}, while simplifying on-device deployment.
\section{Method}
\label{sec:method}

\subsection{Motivation} \label{motivation}
Most speculative decoding approaches exhibit a distinct separation in the training processes of draft and target models. However, directly using an off-the-shelf draft model often leads to sub-optimal performance in many downstream applications. The speculated tokens frequently fail the verification of the target model when draft and target models are misaligned. To achieve improved speedups, fine-tuning both draft and target models on downstream applications becomes necessary. Our objective is to devise an end-to-end trainable single-model framework capable of simultaneously predicting the next token and speculating future tokens. This eliminates the need for an auxiliary draft model while achieving speedups similar to those reported in \cite{leviathan2023fast}. We aim to attain this speedup by increasing the arithmetic intensity of auto-regressive transformer calls without compromising generation quality.

\subsection{Speculative Streaming}

We propose \methodname\xspace to enable parameter efficient speculative fine-tuning and inference of decoder-only models on downstream applications. Compared to the standard draft-target speculative decoding~\cite{leviathan2023fast} and more recently proposed block-wise decoding~\cite{medusa}, \methodname\xspace  introduces the following modifications. (a) Speculative stream design and initialization as described in ~\cref{speculative_finetune} (c) Parallel speculation and verification as described in~\cref{parallel_spec} (d) Parallel tree draft pruning, described in~\cref{tree_pruning} and finally (e) Training objective as described in~\cref{training_objective}.

\subsubsection{Streams Design and Initialization} \label{speculative_finetune}

Parameter efficient fine-tuning \cite{hu2022lora} of decoder-only pre-trained language models involves training low-rank adapters to predict next target token $y_t$ given context tokens $(x_1....x_m)$ and previous target tokens $(y_1 .. y_{< t})$ on downstream applications. To inherently embed a notion of future token planning, we modify the training objective of the target model from next token prediction to n-gram prediction using the concept of multi-stream attention~\cite{qi2020prophetnet,xlnet}. This objective allows the model to plan for future tokens and prevent over-fitting on local correlations. In addition, each of the $\gamma$ streams generates speculative tokens with negligible latency overhead when the model is memory-bound. Specifically, each added stream predicts $p({y_{t + j} | y_{<t}, x)},$ where $1 <= j <= \gamma$, while main stream predicts $p({y_{t} | y_{<t}, x})$. We refer to the multi-head attention mechanism depicted in~\cite{vaswani2017attention} as main-stream self-attention and introduce $\gamma$ additional self-attention streams to speculate future tokens. 

The attention mechanism of main stream is the same as standard multi-head attention \cite{vaswani2017attention}. 

\begin{equation}\label{eq:main_stream_msa}
M_{t}^{k+1} = \text{MHA}(M_{t}^k, M_{\leq t}^k, M_{\leq t}^k)
\end{equation}
Where $M_{t}^k$ denotes hidden states of main stream at layer $k$ and time step $t$ and $MHA(H, H, H)$ denotes attention between query $HW^Q$, key $HW^K$ and value $HW^V$ as described in \cite{vaswani2017attention}.  
On the other hand, each speculative stream $j$ at time step $t$ attends to previous main stream hidden states as well as speculative stream hidden states as: 

\begin{equation}\label{eq:spec_stream_msa}
S_{tj}^{k+1} = \text{MHA}(S_{tj}^k, M_{\leq t}^k \oplus S_{t(\leq j)}^k, M_{\leq t}^k \oplus S_{t(\leq j)}^k)
\end{equation}

where $M_{t}^{k+1}$ and $S_{t}^{k+1}$ refer to main and speculative streams at time step $t$ and layer $k$. Hidden state of last transformer layer $N$, $M_{t}^{N}$ is used to predict $y_t$, whereas each speculative stream at last layer, $S_{tj}^{N}$ predicts  $y_{t+j}$. We refer to layers incorporating the attention mechanism in ~\cref{eq:main_stream_msa} as MHA layers while layers incorporating speculative stream attention ~\cref{eq:spec_stream_msa} are referred to as MSA layers. 
\par
Key/value projections of main stream hidden states are cached during inference to avoid re-computation, whereas, speculative stream attention is specifically designed to avoid storing additional key/value projections associated with individual streams. This is because speculative streams are trained to learn contextual features from main stream key/value context allowing us to not introduce additional caching overhead and operate within memory bounds of resource-constrained devices during inference. We initialize hidden states of speculative streams at layer $N-N_s$ instead of initializing them from the embedding layer, where $N_s < N$. Specifically, stream $j$ at time $t$ is initialized at layer $N-N_s$ as,

\begin{equation}\label{eq:spec_stream_init}
S_{tj}^{N-N_s} = f_{\eta}(M_{t}^{N-Ns}) + P_j^{N-N_s}  
\end{equation}

where $P_j$ is a stream identifier embedding that embeds a sense of relative position into streams and distinguishes the computation from main stream.
$f_{\eta}$ is a linear transformation of rank $\eta$ to transform main stream hidden states into speculative stream hidden states. This initialization helps to reduce computation per forward pass, since only the main stream needs to be passed through $N- N_s$ layers, while speculative streams are passed through the last $N_s$ layers, decreasing the speculative FLOPs contribution by $(N-N_s)/N$ and in turn helping with peak power consumption on the device. In terms of forward pass latency, FLOPs do not contribute significantly when the model is memory bound, however, as we describe in ~\cref{tree_drafting}, we sample additional tokens to make the model compute-bound, therefore FLOP reduction becomes crucial. Initialization with a hidden state of middle-upper transformer layers may also help with the future token prediction as $M^{(N-Ns)}$ itself contains high-level contextual features to aid with the prediction of future n-grams \cite{futurelens}. We also experimented with value rotation based stream design which does not require identifier embeddings and incurs no parameter overhead as described in \ref{value_rotation_sec}.

\subsubsection{Parallel speculation and verification} \label{parallel_spec} \label{tree_drafting}

In standard draft-target speculative decoding~\cite{leviathan2023fast}, speculation and verification processes happen sequentially. The draft model waits for the target model to issue a correction before generating the next draft. The target model also needs to wait for the speculative draft to be generated. \methodname\xspace makes this process more efficient by parallelizing speculation and verification. In each forward pass, the draft generated in the previous step is verified and a new draft is generated as shown in ~\cref{fig:main_fig}. For instance, in step $s$, if draft tokens $(\Tilde{y_1} .. \Tilde{y_{\delta}})$ are accepted where $0 < \delta \leq \gamma$, main stream $M_\delta$ is used to issue a correction token and logits from speculative streams $S_{\delta(1...\gamma)}$ are used to generate draft for step $s+1$.
\par
Instead of using a linear sequence of speculated tokens for verification, we sample a tree of tokens from main and speculative streams, such that each path in the tree is one possible verification candidate. Tree drafting enables accepting the longest matching candidate sequence and more tokens can be advanced during each forward pass. To create a tree draft, instead of sampling 1 token from logits of speculative streams, $(z_1 ... z_\gamma)$, we sample top $k$ tokens and form a tree of sampled tokens as shown in ~\cref{fig:main_fig}, such that tokens sampled from stream $n$ are predecessors of tokens sampled from stream $n+1$. We process a tree draft of speculative tokens in one forward pass by creating an additive attention mask ~\cite{vaswani2017attention} such that each node in the tree attends to its predecessor. Attention mask between $k^{th}$ token sampled from logits of stream $j$, $\Tilde{y}_{jk}$ and the $m^{th}$ token sampled from logits of stream $n$, $\Tilde{y}_{nm}$ is 
\begin{equation} \label{mask_eq} 
    a_{\Tilde{y}_{jk}\Tilde{y}_{nm}} = \begin{cases*}
        0 & if j = n+1, \\
        -\infty & otherwise
    \end{cases*} 
\end{equation}
 Please refer to ~\cref{fig:supp_attn_mask} for more details. It's worth noting that for a fixed $\gamma$ and $k$, the attention mask remains constant in each forward pass and enables effective batching.

\subsubsection{Parallel tree pruning} \label{tree_pruning}

One of the issues with the naive creation of a speculative tree draft is that every permutation between $k$ tokens sampled from each stream needs to be considered as a viable speculative candidate for the next verification pass. For instance, sampling $k$ tokens from each of $\gamma$ streams results in tree draft of size 1 + $\sum_{g=1}^{\gamma}k^g$. Furthermore, each of the draft tokens is batched with $\gamma$ speculative streams in MSA layers to ensure that the generation of the next draft happens in the same forward pass, resulting in a batch size of $(1 + \gamma) * (1 + \sum_{g=1}^{\gamma}k^g)$. As batch size increases, target model inference becomes compute-bound, obviating the latency benefit of sampling more tokens. We mitigate this problem by introducing a parallel tree draft pruning layer, which prunes some of the tokens from the input tree draft based on transition probability between parent and immediate child tokens. 
To obtain transition probabilities without using proxy models, we use an early-exiting-based technique. Specifically, hidden states of the main stream at layer $l$, $M^l$ are passed through a low-rank linear transformation $o_\theta$, where the rank $\theta$ is typically set to a small value like 8 to keep parameter overhead small. We use original language modeling head, $H$ to obtain early exit logits, $\Tilde{z} = H(o_\theta(M^l)$.  $\Tilde{z}_{pc}$ is used to approximate transition probability between parent token $p$ and child token $c$. The pruning layer can be inserted at any point in the network, guided by the trade-off between forward pass latency and pruning accuracy. Early insertion reduces latency but risks pruning potentially valuable tokens. Conversely, late insertion retains more "good" tokens but comes at the cost of increased forward pass latency. In all experiments described in~\cref{sub:expts}, we insert the pruning layer just before speculative stream insertion empirically. More details can be found in ~\cref{fig:supp_imp_extend}.

\subsubsection{Training objective} \label{training_objective}
To efficiently generate future n-grams, we finetune the base model jointly on the prediction loss of the next token as well as $\gamma$ future tokens as 
\begin{align} \label{eq:loss}
L_{ss} =  &-\alpha_0(\sum_{t=1}^{T}\log p_{\theta} (y_t | y_{<t}, x)) \\
& -\sum_{j = 1}^\gamma \alpha_j(\sum_{t=1}^{T-j}\log p_{\theta} (y_{t+j} | y_{<t}, x)) \nonumber
\end{align}
where $\alpha_0$ and $\alpha_j$ are set empirically to normalize losses of the next token and speculative tokens prediction. Tree-pruning adapter described in ~\cref{tree_pruning} can be trained on the next token prediction loss, either jointly along with main and speculative streams or post-training of streams. Training times vary based on the number of MSA layers but are comparable to  \cite{medusa} style approach for $N_s = 4$. For experiments described in \ref{sec:experiments}, our recipe involves training LoRA adapters for 5 epochs on the downstream datasets in  BFloat16, using the AdamQ optimizer, a learning rate of 5e-4, and a linear scheduler. For tree pruning (see \cref{tree_pruning}), we use a low-rank linear transformation of rank 8 to keep parameter overhead minimal.
 
\section{Experiments}
\label{sec:experiments}

\begin{table*}[t] 
\caption{Walltime speedup, CR ratio, parameter overhead, and Metric comparison using different models fine-tuned on downstream applications. CR ratio denotes acceleration agnostic target model call reduction ratio. We use exact match accuracy as a metric for SqlContext, and Rouge1/RougeLSum as a metric for Dialogsum and E2E-NLG tasks. 
}
\label{table:main_results}
\vskip 0.15in
\centering
\begin{tabular}{c|c|ccccc}
\Xhline{1.0pt}  
Dataset & Model & Method & SpeedUp ($\uparrow$) & CR Ratio ($\uparrow$) & Metric ($\uparrow$) & \# Extra Parameters ($\downarrow$)\\
\Xhline{1.0pt}
\multirow{9}{*}{SqlContext} & \multirow{3}{*}{OPT-1.3b} & Baseline & $1.00$  & $1.00$ & $84.98$ & $-$\\
 & & Medusa & $2.07$ & $2.79$ &  $84.98$ & $4.28E8$ \\
 & & SS (ours) & $\mathbf{2.39}$ & $\mathbf{3.57}$ & $\mathbf{87.40}$ & $\underline{4.096E4}$ \\\cline{2-7}
 & \multirow{3}{*}{{PHI-1.3b}} & Baseline & $1.00$ & $1.00$ & $88.71$ & $-$ \\
 & & Medusa & $2.58$  & $3.25$ & $88.71$ & $4.36E8$ \\
 & & SS (ours) & $\mathbf{2.62}$ & $\mathbf{3.53}$ & $\mathbf{89.90}$ & $\underline{4.096E4}$ \\\cline{2-7}
 & \multirow{3}{*}{OpenLlama-7b} & Baseline & $1.00$  & $1.00$ & $89.88$ & $-$\\
 & & Medusa & $\mathbf{3.20}$ & $4.10$ & $90.11$ & $5.91E8$ \\
 & & SS (ours) & $3.14$ & $\mathbf{4.13}$ & $\mathbf{91.70}$ & $\underline{8.19E4}$ \\
 \Xhline{1.0pt}
\multirow{9}{*}{DialogSum} & \multirow{3}{*}{OPT-1.3b} & Baseline & $1.00$ & $1.00$ & $43.40$/$35.56$ & $-$ \\
 & & Medusa & $1.56$  & $1.91$ & $43.40$/$35.50$ & $4.28E8$ \\
 & & SS (ours) & $\mathbf{1.94}$ & $\mathbf{2.62}$ & $\mathbf{44.07}$/$\mathbf{35.99}$ &  $\underline{4.096E4}$ \\\cline{2-7}
& \multirow{3}{*}{PHI-1.3b} & Baseline & $1.00$  & $1.00$ & $\mathbf{43.57}$/$\mathbf{35.60}$ & $-$\\
& & Medusa & $\mathbf{1.89}$ & $2.28$ & $\mathbf{43.57}$/$\mathbf{35.60}$ &  $4.36E8$ \\
& & SS (ours) & $1.83$ & $\mathbf{2.34}$ & $43.36$/$35.31$ & $\underline{4.096E4}$ \\\cline{2-7}
& \multirow{3}{*}{OpenLlama-7b} & Baseline & $1.00$  & $1.00$ & $\mathbf{44.20}$/$\mathbf{36.50}$ & $-$ \\
 & & Medusa & $1.76$  & $2.25$ & $\mathbf{44.20}$/$\mathbf{36.50}$ & $5.91E8$ \\
 & & SS (ours) & $\mathbf{1.87}$ & $2.51$ & $43.92$/$35.70$ & $\underline{8.19E4}$ \\
\Xhline{1.0pt}
\multirow{9}{*}{E2E-NLG}  & \multirow{3}{*}{OPT-1.3b} & Baseline  & $1.00$ & $1.00$ & $\mathbf{69.48}$/$50.17$ & $-$ \\
 & & Medusa & $2.13$ & $2.95$ & $\mathbf{69.48}$/$50.17$ & $4.28E8$ \\
 & & SS (ours) & $\mathbf{2.45}$ & $\mathbf{3.72}$ & $69.32$/$\mathbf{50.51}$ & $\underline{4.096E4}$ \\\cline{2-7}
 & \multirow{3}{*}{PHI-1.3b} & Baseline  & $1.00$ & $1.00$ & $\mathbf{67.90}$/$48.50$ & $-$ \\
 & & Medusa & $2.78$ & $3.35$ & $\mathbf{67.90}$/$48.50$ & $4.36E8$ \\
 & & SS (ours) & $\mathbf{2.84}$ & $\mathbf{3.69}$ & $67.40$/$\mathbf{48.52}$ & $\underline{4.096E4}$ \\\cline{2-7}
 & \multirow{3}{*}{OpenLlama-7b} & Baseline & $1.00$  & $1.00$ & $\mathbf{69.50}$/$\mathbf{50.30}$ & $-$ \\
 & & Medusa & $2.70$ & $3.22$ & $\mathbf{69.50}$/$\mathbf{50.30}$ &  $5.91E8$ \\
 & & SS (ours) & $\mathbf{2.96}$ & $\mathbf{3.55}$ & $68.66$/$49.56$ &  $\underline{8.19E4}$ \\
\Xhline{1.0pt}
\end{tabular}
\end{table*}

\begin{table*}[t] 
\caption{Walltime latency (per sample) comparison with standard draft-target based speculative decoding approach using OPT-125m as the draft model for $\gamma = 4$. Although calls to target model using our approach are higher than draft-model-based speculative decoding, it does not incur auto-regressive drafting overhead, achieving better latency on OPT-1.3b and OPT-6.7b models. We use exact match accuracy as a metric for SqlContext, while Rouge1/RougeLSum is used as a metric for Dialogsum and E2E-NLG tasks.
}
\label{table:two_stage}
\vskip 0.15in
\centering
\begin{adjustbox}{width=\textwidth}
\begin{tabular}{c|c|ccccc}
\Xhline{1.0pt}  
Dataset & Target & Method & Target calls & Draft Calls & Walltime Latency ($ms$, $\downarrow$) & Metric ($\uparrow$) \\
\Xhline{1.0pt}
\multirow{4}{*}{SqlContext} & \multirow{2}{*}{OPT-1.3b} & Two-model SD & $6.59$ & $22.35$ & $269.24$ & $84.98$ \\
 & & SS (ours) & $7.79$ & $0$ & $\mathbf{133.48}$ & $\mathbf{87.40}$ \\\cline{2-7}
 &  \multirow{2}{*}{OPT-6.7b} & Two-model SD &  $6.60$ & $22.41$ & $301.10$ & $89.13$ \\
 & & SS (ours) & $6.88$ & $0$ & $\mathbf{157.04}$ & $\mathbf{89.34}$  \\
\hline
\multirow{4}{*}{Dialogsum} & \multirow{2}{*}{OPT-1.3b} & Two-model SD & $11.65$ & $42.59$ & $493.59$ & $43.40$/$35.60$\\
 & & SS (ours) & $13.41$ & $0$ & $\mathbf{248.26}$ & $\mathbf{44.07}$/$\mathbf{35.99}$ \\\cline{2-7}
   & \multirow{2}{*}{OPT-6.7b} & Two-model SD & $12.15$ & $35.76$ & $555.99$ & $\mathbf{44.40}$/$\mathbf{36.60}$ \\
  & & SS (ours) & $14.39$ & $0$ & $\mathbf{442.83}$ & $44.30$/$36.30$ \\
\hline
\multirow{4}{*}{E2E-NLG} & \multirow{2}{*}{OPT-1.3b} & Two-model SD & $8.86$  & $31.47$ & $345.72$ & $\mathbf{69.48}$/$50.17$ \\
 & & SS (ours)  & $9.80$ & $0$  & $\mathbf{164.23}$ & $69.32$/$\mathbf{50.51}$   \\\cline{2-7}
  & \multirow{2}{*}{OPT-6.7b} & Two-model SD &  $8.90$ & $31.58$ & $412.02$ & $\mathbf{69.34}$/$\mathbf{49.88}$ \\
  & & SS (ours) & $10.26$ & $0$ & $\mathbf{243.62}$ & $69.07$/$49.69$ \\
\hline
\Xhline{1.0pt}
\end{tabular}
\end{adjustbox}

\end{table*}

We evaluate our methods on the pre-trained models of various scales and a diverse set of downstream applications. 

\textbf{Dataset.} We test our methods on a diverse set of applications that are vital to on-device AI assistants, namely Structured Queries, Text Summarization, and Meaning Representation. We specifically choose fine-tuning setup since it has been a norm to share a base model and use application-specific adapters for user-facing applications. We use the Dialogsum~\cite{chen-etal-2021-dialogsum} dataset for Text Summarization, the sql-create-context dataset built from WikiSQL~\cite{zhongSeq2SQL2017} and SPIDER~\cite{yu2018spider} for Structured Queries, and e2e-nlg dataset~\cite{dusek.etal2020:csl} for Meaning Representation.

\textbf{Model Configuration.} We tested four different open source models of various scales, Phi(1.3B)\cite{textbooks2}, Open-llama(7B)\cite{touvron2023llama}, and OPT(1.3B, 6.7B) \cite{zhang2022opt}. We compare our method with the standard draft-target speculative decoding (\cite{leviathan2023fast}) and single-model speculative decoding framework, Medusa~\cite{medusa}. For the standard draft-target approach, we use OPT-125m, the smallest configuration of available open-source OPT models as the draft model.

\textbf{Baselines} 
To compare with Medusa~\cite{medusa} style approach, we use pre-trained base models with LoRA adapters~\cite{hu2022lora} of rank 32 and Medusa heads as the baseline, and \methodname\xspace with the same base models, stream embeddings and LoRA adapters as target. Medusa heads are trained following the recipe described in \cite{medusa}. Both Medusa heads and the number of maximum streams are fixed to 4 and the residual blocks per head used in Medusa are set to 1. For comparison with standard draft-target speculative decoding \cite{leviathan2023fast}, we use OPT models since they come with different configurations and sizes. OPT-125m is deployed as a draft model while OPT-1.3b and OPT-6.7b are used as target models since a ratio of 10-100X is typically considered to be optimal. Note that, similar to the target model, only LoRA adapters of the draft model are fine-tuned on downstream applications because fine-tuning the entire draft model on each downstream application is not practical in on-device settings. Also, LoRA fine-tuning tends to achieve on-par performance to full-model fine-tuning~\cite{hu2022lora}. 

\subsection{Results} \label{sub:expts}

\subsubsection{Overview}
We report wall-time speedups and generation quality metrics on test split using a batch size of 1 on a single Nvidia A100-80G GPU. Inference is performed in float16 using greedy sampling and $T = 0$. Please refer to ~\cref{supp_expt_details} for more experimental details and ~\cref{supp_ablations} for ablations on top-k sampling and $T = 1$. We use  Exact Match (EM) accuracy metric for the structured query task and Rouge1/RougeLSum metrics for the Dialog Summarization and Meaning Representation tasks. We use $N_s/N$ of $1/6$ for the structured query task and $1/2$ for the summarization and meaning representation task. $N_s$ is chosen to ensure the generation metric is on-par with the baseline. Details on the effect of $N_s$ on generation metric are found in ~\cref{n_layers_ablation}. 
\par
~\cref{table:main_results} presents the comparison between standard auto-regressive decoding baseline, Medusa, and our approach in terms of speedup, call reduction ratios, and the number of extra parameters. We find that across a variety of downstream tasks, the walltime speedups and call reduction ratios of \methodname\xspace are consistently on-par/higher than Medusa while incurring significantly lesser parameter overhead. Furthermore, as summarized in \cref{table:two_stage}, our approach achieves better wall-time latencies than the standard draft-target speculative decoding since the difference in the number of target calls between both approaches is not large enough to offset auto-regressive drafting overhead. All wall-time latencies are reported using open-source versions of models available on \cite{wolf2019huggingface} and it is possible that further optimizing draft and target models using efficient inference techniques \cite{nvidiafastertransformer} or quantization (int4/8) may lead to lower latencies. Finally, It's worth noting that the generation metrics of our method are consistently comparable with LoRA fine-tuned base models making it an excellent alternative to next-token prediction-based fine-tuning.  

\subsubsection{Analysis and Insights} \label{sub: speedup_anal}
\textbf{Without Auxiliary Models } Medusa heads generate each token independently from the shared hidden state of the last layer, and dependency between speculative tokens predicted by medusa heads, $y_{(t+1 .. t+\gamma)}$ and next token $y_{t}$ predicted by the base model at time step $t$ may not be well captured since there no attention mechanism involved. On the other hand, speculative streams attend to the main stream and each other, capturing the token dependency, resulting in better call reduction ratios than Medusa. 
In terms of parameters, each Medusa head adds about $h^2 + hv$ parameters, where $h$ is the hidden size and $v$ is the vocabulary size. The number of Medusa heads also scales linearly \wrt $\gamma$, the length of the speculative window, which in turn increases parameter overhead linearly with $\gamma$. On the other hand, \methodname\xspace uses speculative adapters which do not scale with $\gamma$. Although, Stream identifier embeddings scale with $\gamma$, the parameter overhead associated with each embedding is linear to $h$. Furthermore, in fine-tuning settings ``speculative adapter" parameters are shared with base model adapters, therefore, parameter overhead associated with our approach is just $\gamma h$. 

\begin{figure}
    \centering
    \includegraphics[width=0.9\linewidth]{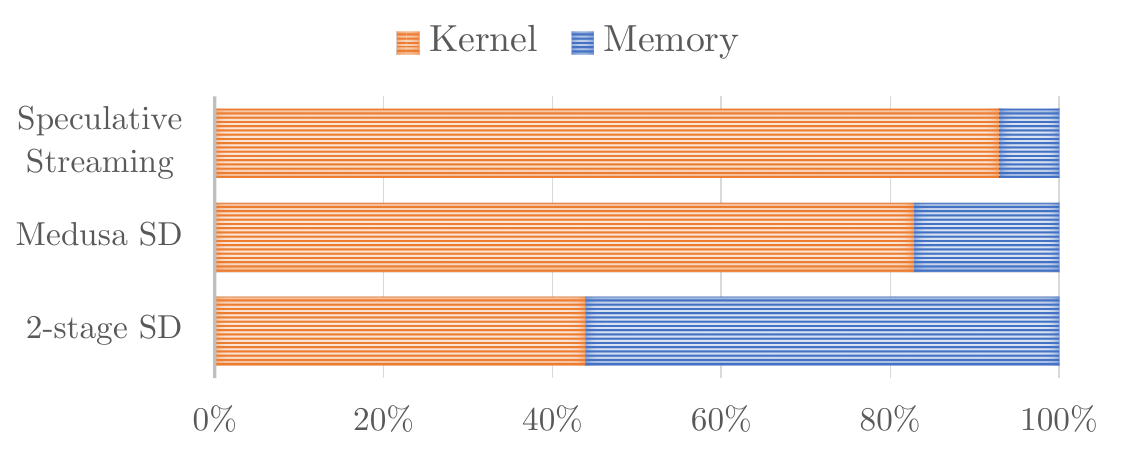}
    \caption{\methodname\xspace speeds up decoding by increasing arithmetic intensity of memory bound auto-regressive decoding step. Kernel and memory utilization of OPT-1.3b model with Medusa-style approach and draft model (OPT-125m) based speculative decoding approach is also shown for comparison. 
    }
    \label{fig:kernel_memory}
\end{figure}

\textbf{With Auxiliary Models } \methodname\xspace consistently achieves lower wall time latency than standard draft-target speculative decoding as depicted in \cref{table:two_stage}. It’s worth noting that,  target model calls of draft-target speculative decoding are lower than \methodname\xspace, however, it comes at the cost of auto-regressively running draft model $\gamma$  times to generate speculative draft.  On the other hand, draft generation with \methodname\xspace  incurs almost no additional latency overhead, as target model decoding tends to be memory-bound even with increased tree draft size. This translates to increased kernel utilization and arithmetic intensity as shown in \cref{fig:kernel_memory}. Draft-based approach on the other hand has low kernel utilization because of the memory-bound nature of auto-regressive drafting.
\par
An argument could be made that a smaller draft model may perform better since drafting should cost less, but acceptance rates may drop as well as the draft model size is decreased. To formalize the comparison with standard draft-target speculative decoding, we do the following analysis, let's say, $C_{draft}$ is the latency cost associated with forward pass through the draft model, $C_{target}$ is the cost associated with forward pass through target model, while $C_{ss}$ is cost associated with speculative streaming forward pass. $\zeta$ is the number of decoding tokens advanced during the verification step for the draft-target approach while $\beta$ is the number of tokens advanced in \methodname\xspace.  We equate latency cost associated with single token advancement to compare both approaches.

\begin{align}
 &(\gamma*C_{draft} + C_{target})/\zeta = C_{ss} / \beta \\
 (\gamma + &C_{target}/C_{draft})/\zeta = (C_{ss}/C_{draft}) / \beta \nonumber
\end{align} 

\begin{figure}
    \centering
    \includegraphics[width=0.7\linewidth]{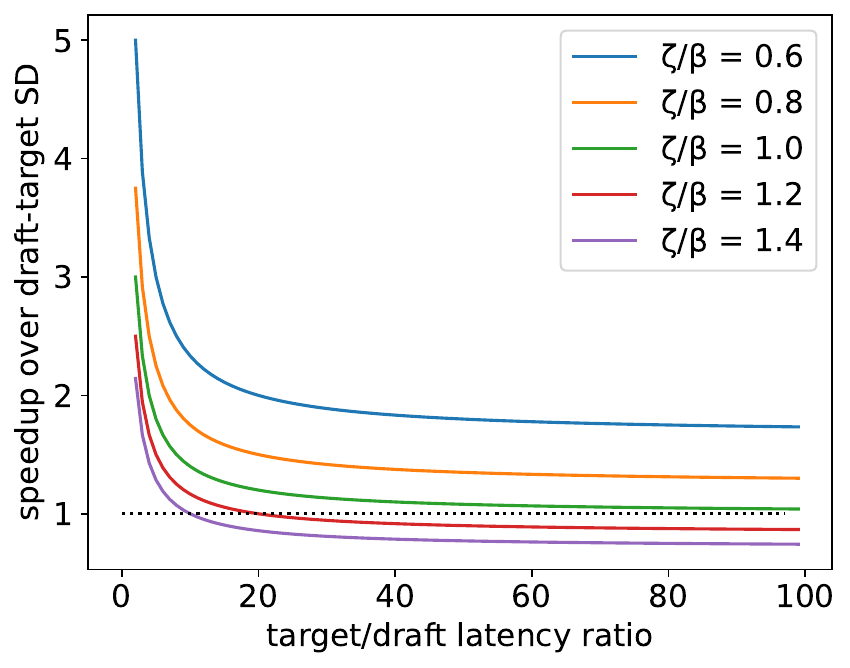}
    \caption{\methodname\xspace speedup over draft-based speculative decoding for different $\zeta/\beta$ and target/draft latency ratios, where $\zeta$ denotes the number of advancements per verification step for draft-based speculative decoding while $\beta$ denotes the same for \methodname. }
    \label{fig:speedup_over_draft_based}
\end{figure}

Assuming $\gamma = 4, C_{target}/ C_{draft} = 10$, and $C_{ss} \approx C_{target} $, $\zeta = 1.4\beta$, meaning that advancements per verification step in standard draft-target approach have to be 1.4X of \methodname\xspace to achieve wall time latency parity. Note that, this analysis ignores cache adjustment overhead and prompt processing overhead,  but provides valuable intuition to guide the choice between draft-target vs \methodname\xspace approaches. We also analyze under which settings speculative streaming is likely to offer more benefits as compared to the standard draft-target approach.  Fig. \ref{fig:speedup_over_draft_based} shows theoretical speedups of \methodname\xspace over draft-target based approach for different Target to draft latency ratios. As the latency ratio increases, the draft-target approach is likely to offer more speedup benefits when $\zeta/\beta > 1$, meaning that when the draft model is accurate enough to achieve more token advancements per target model verification step than \methodname\xspace and also small enough to yield higher latency ratios, it is likely to benefit more. Finding/creating such a model usually requires significant engineering efforts. In downstream application settings, finding ideal draft models becomes even more challenging since $\zeta$ tends to vary based on application. If applications share the draft model and only train adapters, the draft model may not remain small enough to meet target-to-draft latency ratios, making it challenging to achieve more speedups than Speculative Streaming. 

\begin{figure}
    \centering
    \includegraphics[width=0.9\linewidth]{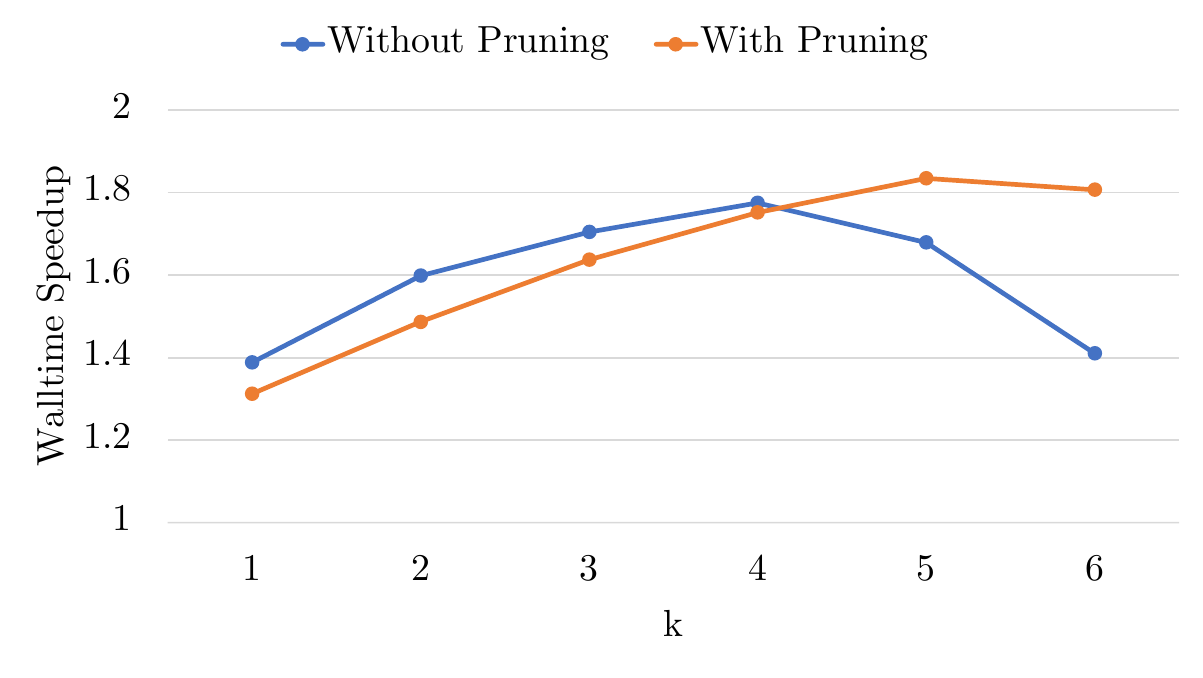}
    \caption{As more tokens ($k$) are sampled from each stream keeping $\gamma$ fixed for the creation of a tree draft, walltime speedup increases due to the increased number of candidates. This trend reverses as $k$ continues to increase and the model transits into the compute-bound phase. Pruning less probable paths from tree draft helps to reduce compute for higher values of $k$ thereby reducing latency per forward pass and offering more speedup. 
    }
    \label{opt_ablation}
\end{figure}

\subsection{Ablations}

\textbf{Speculative Draft Size.}
To improve the acceptance rate of the tree draft, we try various settings of $\gamma$, the number of speculative positions, and $k$, the number of sampled tokens per speculative position. \cref{opt_ablation} shows walltime speedup for $\gamma = 3$. As we sample more tokens from each speculative position, advancement per forward pass, $\beta$ increases since more candidates are available for verification, leading to more speedup. However, as we continue to increase $k$, forward pass latency overhead becomes more prevalent as the model transitions into compute-bound phase and the speedup reverses the course. This is because naively forming a tree draft leads to an exponential increase in batch size with $k$ as described in \ref{tree_pruning}. We insert a tree pruning layer to remove less probable paths and reduce the size of the tree draft. Pruning tree draft reduces forward pass latency, and a well calibrated threshold ensures that only noisy paths in the tree get pruned. Tree pruning tends to help with walltime speedup as $k$ continues to increase as shown in \cref{opt_ablation}.

\begin{figure}
    \centering
    \includegraphics[width=0.9\linewidth]{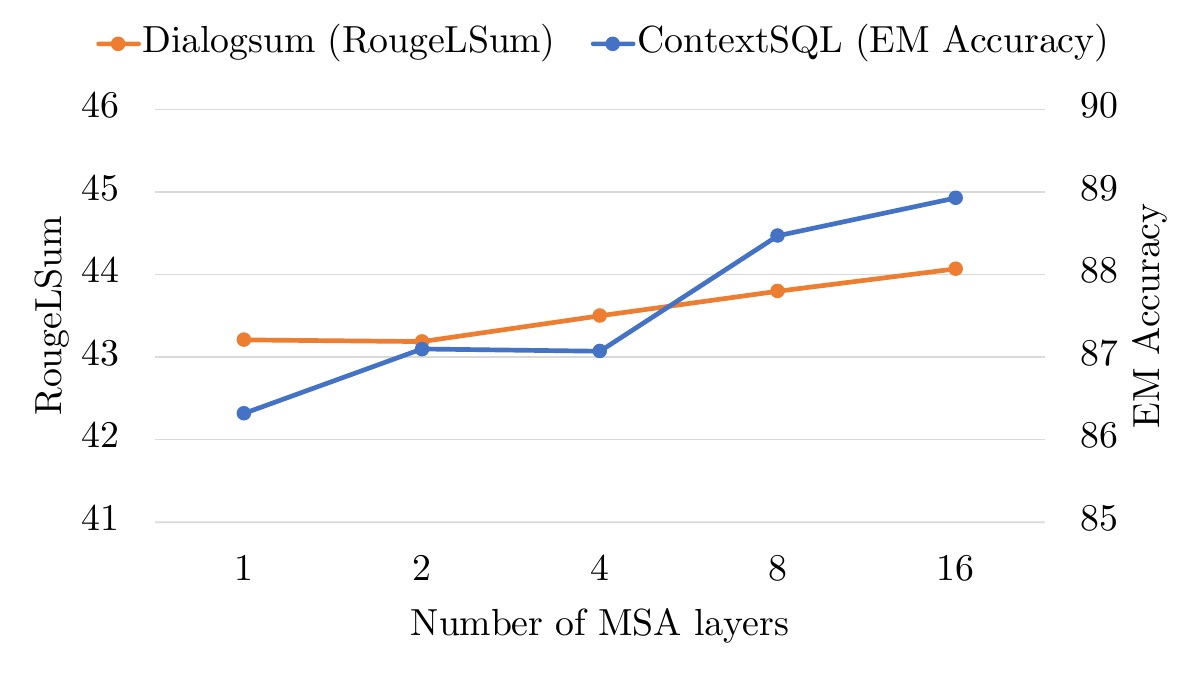}
    \caption{As the number of multi-stream attention layers increases, metrics on downstream tasks improve as well. We use RougeLSum as the metric for the Dialogsum task, and Exact Match (EM) accuracy as the metric for the ContextSQL task. 
    }
    \label{fig:n_layers_ablation}
\end{figure}

\textbf{Number of MSA Layers} \label{n_layers_ablation}
There are trade-offs involved in deciding the number of MSA layers to incorporate in terms of downstream generation metric, training time, and FLOPs increase. As we increase the number of MSA layers, the generation metric improves and this trend remains the same across different downstream tasks. 
Typically incorporating MSA in the top 2 - 8 layers offers a good trade-off between metric,  FLOPs increase and training time. Figure \ref{fig:n_layers_ablation} shows the generation performance of the OPT-1.3b model on Structured Query and Summarization tasks.

\section{Conclusion}
\label{sec:conclusion}
In this paper, we proposed \methodname\xspace, a method to accelerate decoding of large language models. Compared to the standard speculative decoding approaches, \methodname\xspace removes the need for an auxiliary ``draft'' model. Instead, it unifies speculation and verification by efficiently fusing multiple speculative streams into a single ``target'' model. \methodname\xspace simplifies the fine-tuning process and achieves  on-par or better speed-up and quality compared to previous approaches. It is also parameter efficient and removes the need for loading two models into the memory, making it a suitable approach for resource-constrained scenarios.

\section*{Acknowledgements}
We would like to thank
Sachin Mehta, Moin Nabi, Antonie Lin, Minsik Cho, Arsalan Farooq, and Jason  Williams
for their valuable feedback and discussions.

\bibliography{speculative_decoding}
\bibliographystyle{icml2024}

\newpage
\appendix
\onecolumn

\section{Implementation Details}
\subsection{Tree Draft Management}

\begin{figure}[!hb]
    \centering
    \includegraphics[width=1.1\linewidth]{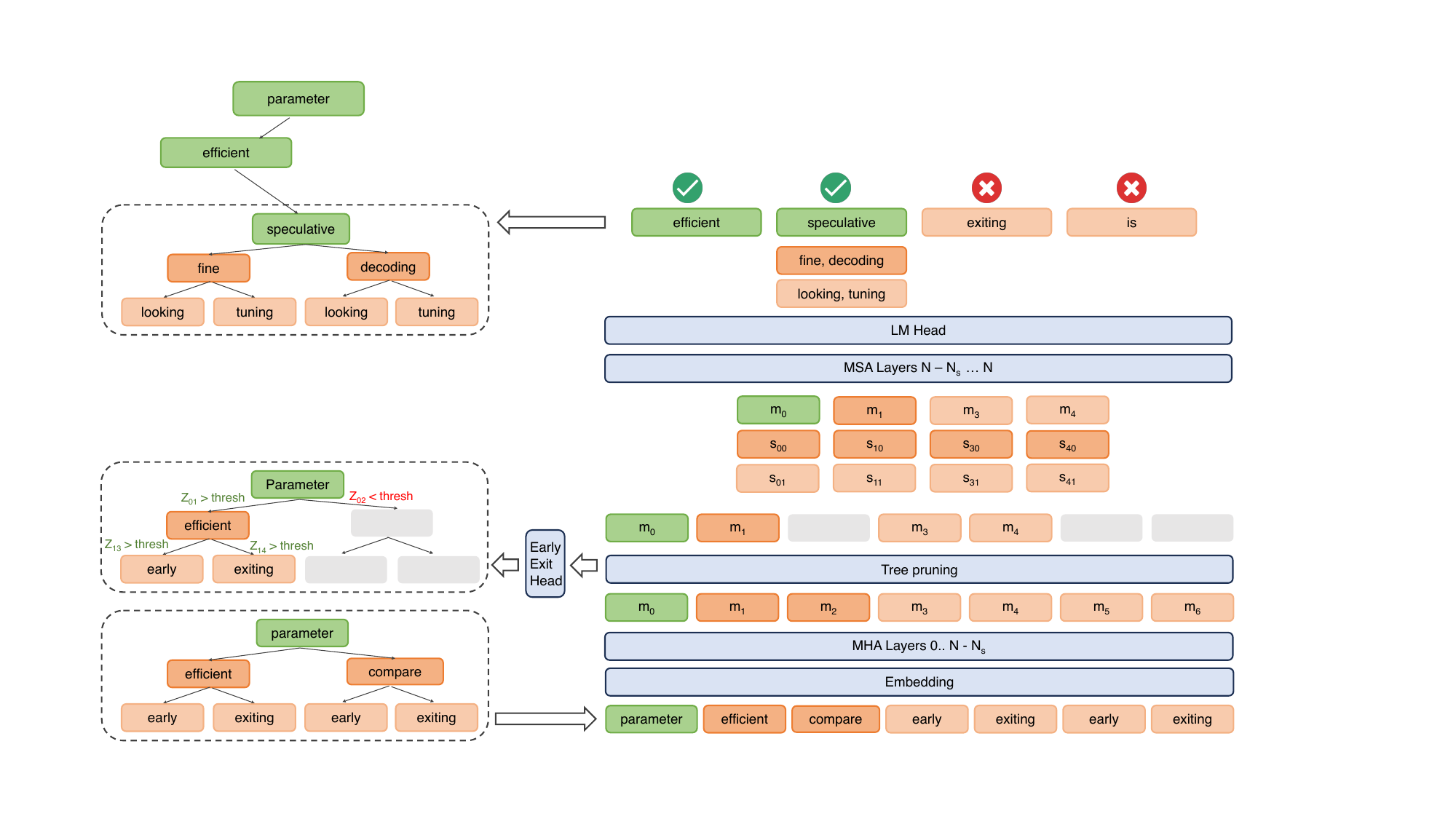}
    \caption{Parallel tree draft speculation and verification: Tree draft from the previous iteration is flattened for verification. After $N - N_s$ MHA layers, the tree pruning procedure obviates less probable tokens based on transition probability between parent and child tokens. In this illustration $Z_{i}$ denotes normalized early exit logits corresponding to main stream at index $i$, $m_i$, while $Z_{ij}$ denotes transition probability between token at index $i$ and $j$ in flattened tree draft. The verification procedure is subsequently run on the pruned tree and speculative tokens are sampled from streams corresponding to the latest accepted token. In above illustration, $``speculative''$, $``fine, decoding''$ and $``looking, tuning''$ are sampled from streams $m_1$, $s_{10}$ and $s_{11}$.}
    \label{fig:supp_imp_extend}
\end{figure}

\begin{figure}
    \centering
    \includegraphics[width=0.3\linewidth]{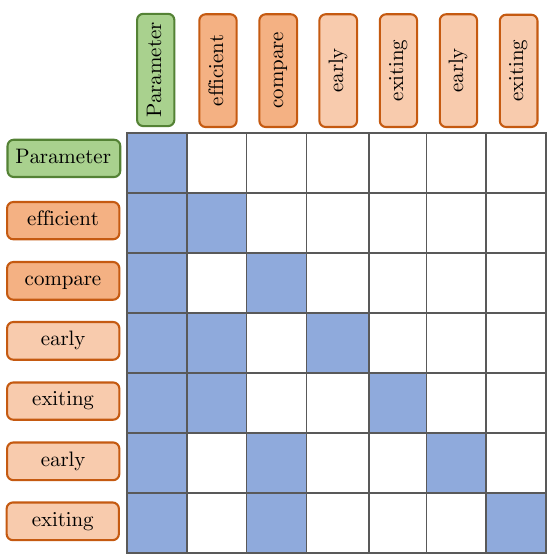}
    \caption{Attention mask for tree draft is composed in such a way that child tokens can attend to all predecessors starting from root, root being correction issued by main stream. In this illustration, $``early``$ attends to $``parameter``$ and $``efficient``$ and itself since $``parameter - efficient - early ``$ forms one path in tree. $``early``$ is also replicated to form another path $``parameter - compare - early ``$. This attention mask allows batching multiple paths and increasing acceptance rate as number of candidates increase. }
    \label{fig:supp_attn_mask}
\end{figure}

In this section, we go into more detail about tree draft sampling, flattening, and pruning. As shown in the main paper, when processing prompt $(x_1...x_t)$, we insert speculative streams along with the last token to generate logits, $z_t$ corresponding to main stream and $(z_{t1} ... z_{t\gamma})$ corresponding to speculative streams. Tree draft is sampled following the procedure described in ~\cref{parallel_spec}. The sampled draft is then flattened along the sequence length dimension and the attention mask is composed such that child nodes attend to their predecessors starting with root as shown in ~\cref{fig:supp_imp_extend} and ~\cref{fig:supp_attn_mask}. The root token of the tree draft is the correction issued by main stream. Each iteration after prompt processing involves verifying the previous tree draft and sampling a new one. After passing the tree draft through $N-N_s$ layers, we use contextual features learned by middle layers to approximate transition probability between parent and child tokens.  As shown in  ~\cref{fig:supp_imp_extend}, since the transition probability between token $``parameter''$ and $``compare''$ is less than a set threshold, we prune the sub-tree starting from $``compare"$ in the feature domain , and $m_2, m_5, m_6$ are pruned. Please note that the key value cache of layers $0 .. (N-N_s-1)$ before the pruning layer is not trimmed at this point to keep pruning latency overhead minimal. Key value cache backtracking is done lazily after each generation step. Speculative streams are inserted alongside each node in the pruned draft. Layers $(N-N_s .. N)$ use Multi-stream attention as described in ~\cref{eq:spec_stream_msa}. The verification procedure finds the longest matching path in the pruned tree that main stream can accept. As shown in ~\cref{fig:supp_imp_extend}, path $(``parameter'', ``efficient'', ``speculative'')$ is accepted. Correction token sampled from logits of main stream corresponding to last accepted token, $m_1$ becomes new root while tokens sampled from logits of streams $(s_{10}, s_{11})$ form the sub-tree. 

\subsection{Experimental Setup Details} \label{supp_expt_details}

Fine-tuning procedure for both baseline and target approaches described in the paper in ~\cref{sec:experiments}  involves training LoRa adapters for 5 epochs. We set $\alpha_0 = 1$ and $\alpha_j = 0.1$ for $j=1...\gamma$ to weigh speculative loss relative to next token prediction loss for both baseline and target methods. We experimented with linear transformations of different ranks to initialize speculative streams from main stream as described in ~\cref{eq:spec_stream_init}, however we find that simply using identity transformation achieves similar performance with much less parameter overhead. We use identity transformation for all the experiments described in~\cref{sec:experiments}.
We report best results for Medusa and our approach over different $\gamma$ and $k$ values, while for standard draft-target speculative decoding approach $k$ is fixed to 1. We also report accelerator agnostic speedups (call reduction ratios) assuming negligible verification and draft composition overhead as latency of forward pass, verification and draft composition procedures vary greatly depending on accelerator (\eg a mobile device neural engine \vs Nvidia A100), while call reduction ratio metric tends to serve as roof-line for achievable speedup. Lastly, we use ``hard`` matching criteria for verification of speculative draft. Relaxing this criteria to ``soft`` matching may yield higher speedups \cite{medusa}

\begin{figure}[!ht]
    \centering
    \includegraphics[width=\linewidth]{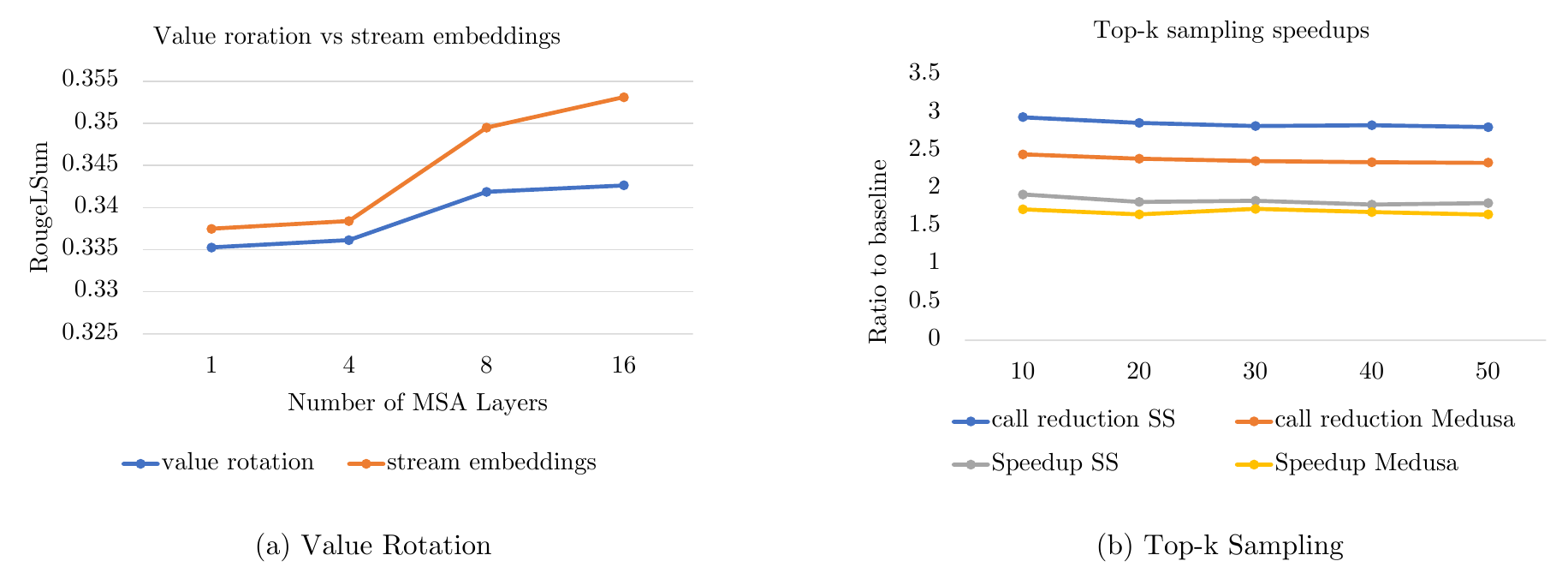}
    \caption{(a) We analyze the effect of value projection rotation on RougeLSum scores of the Dialog summarization task using PHI-1.3b as the base model for different numbers of MSA layers. Each stream is rotated in proportion to the distance from the main stream. (b) We study the effect of top-k sampling on wall-time speedups and call reduction ratios for \methodname\xspace (SS) and Medusa style approaches using OPT-1.3b as a base model on the Meaning Representation task.}
    \label{fig:supp_ablations}
\end{figure}

\section{Ablation: } \label {supp_ablations}

\textbf{Value Rotation} \label{value_rotation_sec}
We analyzed more ways of differing computation of main stream from speculative streams. Apart from using dedicated stream embeddings, one way to differentiate the computation while incorporating a sense of relative position is simply rotating streams relative to each other. In this ablation, we initialize each stream with the main stream hidden state and rotate the value projection during attention computation in the proportion of the relative distance from main stream as : 

\begin{equation}
    V^k_{tn} = V^k_te^{i\epsilon n}  
\end{equation}

Where $1 <= n <= \gamma$ is stream index, $V^k_{t}$ denotes value projection of main stream at time step $t$ and layer $k$, while $V^k_{tn}$ denotes value projection of stream n, $ 0 \leq \epsilon \leq \frac{\pi}{2N}$ denotes an arbitrary rotation step and $N$ denotes the sum of maximum sequence length and number of streams. \cref{fig:supp_ablations} (a) shows the effect of using value rotation on Rouge scores on the Dialog Summarization task with the Phi-1.3b model. Downstream metric for value rotation-based approach tends to be lower than using dedicated stream embeddings across different settings of MSA layers, however, the trend of increasing metric with added MSA layers remains the same. It's worth noting that for $N_s = 16$, simply rotating value projections achieve better metrics than using $N_s = 4$ with dedicated stream embeddings.  

\textbf{Top-k Sampling}
In the main paper, we reported speedup results using greedy sampling and T=0. To further analyze speedups in the Top-k sampling regime, we try various values of $k$ and T = 1 for both Medusa style and \methodname\xspace approaches. \cref{fig:supp_ablations} (b) shows the effect of increasing $k$ on the walltime speedups and call reduction ratios. Although increasing $k$ leads to lower wall-time speedups for both baseline and target methods due to stochastic rejection of tokens, our approach retains its lead achieving better call reduction ratios and walltime speedups across different values of $k$.

\section{Compute and Memory Profiling}

\begin{figure*}
    \centering
    \includegraphics[width=\linewidth]{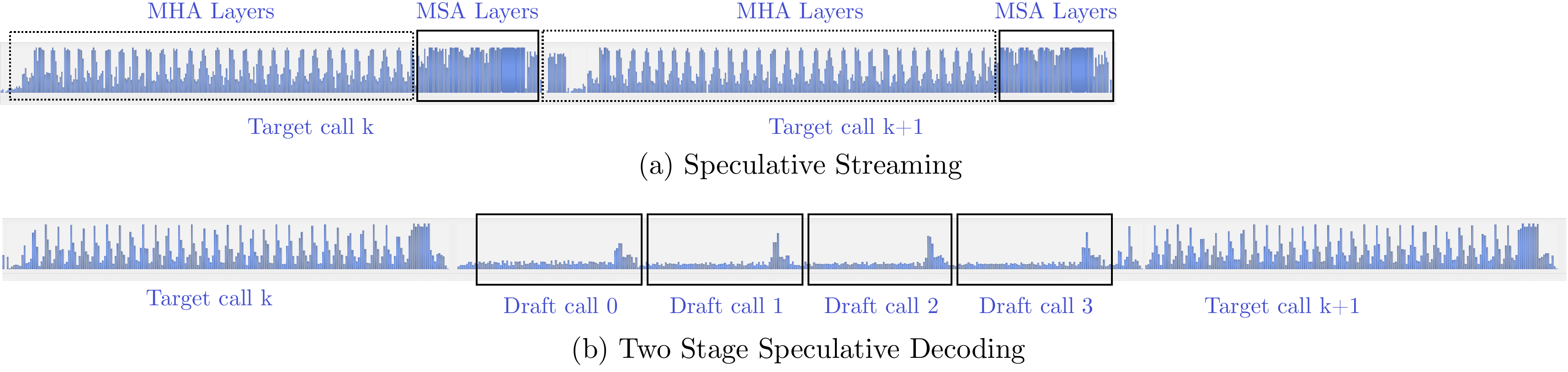}
    \caption{Kernel utilization timeline for speculative streaming and the standard draft-target speculative decoding. Draft-target approach runs speculation and verification in serial manner while it is parallelized in \methodname. Auto-regressive draft generation often has low kernel utilization as shown leading to decreased overall kernel utilization while MSA layers in \methodname \  increase kernel utilization by generating a non-autoregressive draft and speeding up decoding significantly.}
    \label{fig:supp_profiling}
\end{figure*}

The draft overhead associated with the standard draft-target speculative decoding approach tends to be non-trivial especially when the latency ratio between target and draft models $c_{target}/c_{draft} <= 10$. This is because speculation and verification procedures are run in serial manner. \cref{fig:supp_profiling} shows the kernel utilization timeline when OPT-125m is used as a draft while OPT-1.3b model is used as the target. Auto-regressive draft generation decreases overall kernel utilization in draft-target approach, while additional computation involved in MSA layers increase kernel utilization in case of \methodname\xspace thereby efficiently utilizing the accelerator and speeding up the decoding process. Negligible cost draft models may offer a better choice to keep kernel utilization at higher levels in case of draft-target approach, however, acceptance rates tend to drop as draft model size  decreases. 

\section{Qualitative Examples}

\begin{figure}
    \centering
    \includegraphics[width=1\linewidth]{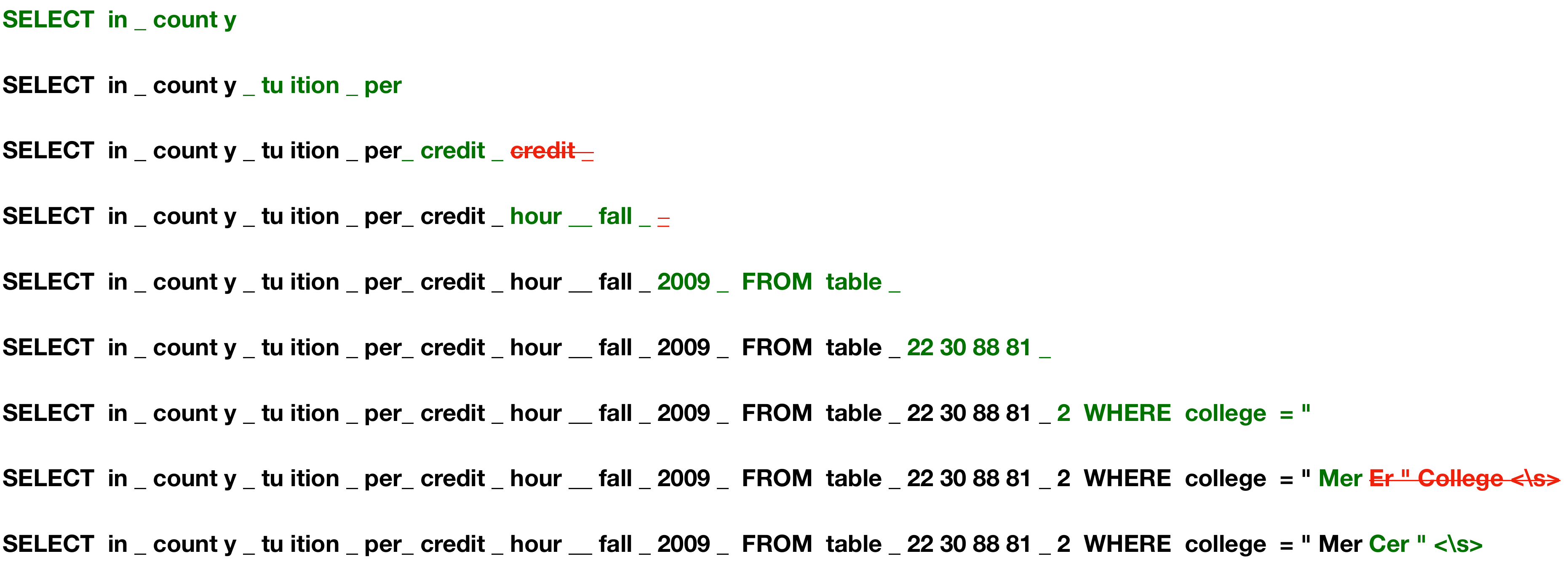}
    \caption{Speculative streaming on SQL generation task for $\gamma = 4$ and $k = 1$, each pass verifies the previous draft and generates a maximum of 5 tokens. For instance in pass 4, $``credit"$ and $``\_"$ (shown in red) are rejected and $``hour" , ``_\_" , ``fall", ``\_" , ``\_"$ are speculated.}
    \label{fig:sql_gen}
\end{figure}

\begin{figure}
    \centering
    \includegraphics[width=1\linewidth]{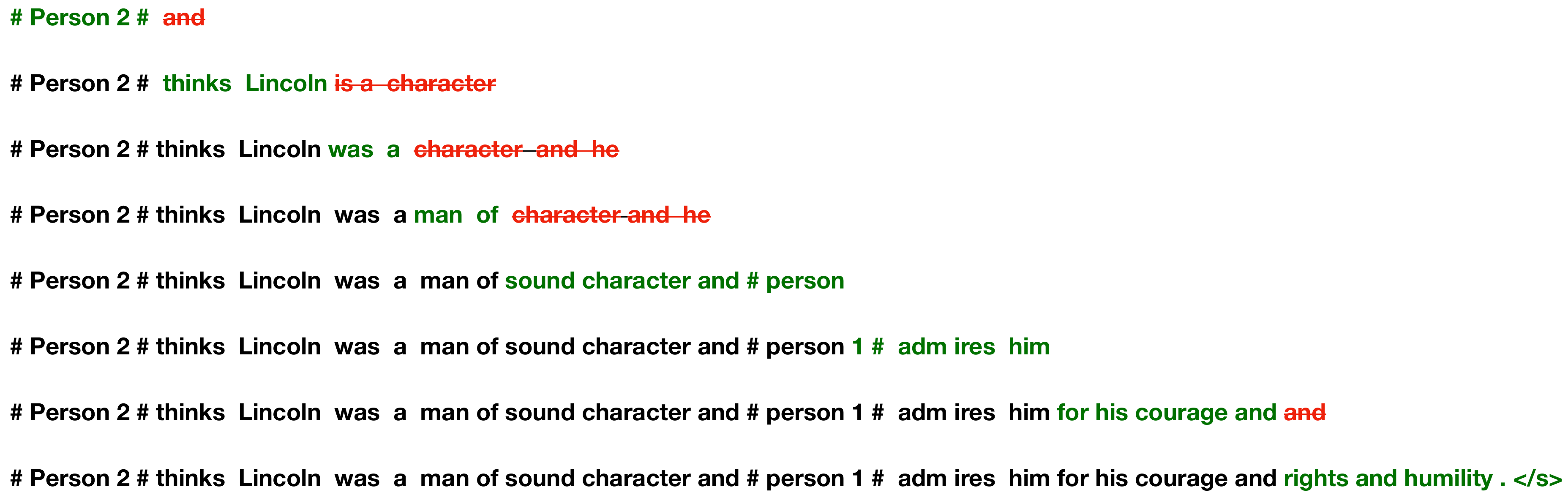}
    \caption{Speculative streaming on Dialog Summarization task for $\gamma = 4$ and $k = 1$, each pass verifies the previous draft and generates a maximum of 5 tokens. For instance, in pass 3, $``is", ``a", ``character"$ are rejected and $``was", ``a", ``character", ``and", ``he"$ are speculated.}
    \label{fig:sumar_gen}
\end{figure}

In this section, we present qualitative examples to illustrate the effectiveness of \methodname\xspace. By examining specific instances, we aim to highlight how this approach enhances the overall performance of the decoding process. An example of the SQL query generation task is shown in Figure~\ref{fig:sql_gen}, while a dialog summarization example is shown in Figure~\ref{fig:sumar_gen}. Each row indicates the previous sequence of accepted draft tokens (in black) and the new sequence of generated tokens in green/red. We use $\gamma = 4$ and $k = 1$ to illustrate the decoding process. Green tokens in each row indicate tokens accepted in the next forward pass, while red tokens indicate tokens rejected in the next forward pass. \methodname\xspace appears to generate meaningful drafts with high acceptance rates by capturing dependencies between tokens quite effectively, despite generating them in a non-auto-regressive manner.

\end{document}